%% file: bmvc_paper.tex
\documentclass{bmvc2k}

\title{Continuous Levels of Detail for Light Field Networks}

\addauthor{David Li}{https://davidl.me}{1}
\addauthor{Brandon Y. Feng}{https://brandonyfeng.github.io}{1}
\addauthor{Amitabh Varshney}{https://www.cs.umd.edu/~varshney/}{1}

\addinstitution{
 University of Maryland, College Park\\
 Maryland, USA
}

\runninghead{Li, Feng, Varshney}{Continuous Levels of Detail for LFNs}

\def\etal{\emph{et al}\bmvaOneDot}

\input{text/00_Macros}

\usepackage{float}
\usepackage{caption}
\usepackage{subcaption}
\usepackage{amsmath}
\usepackage[linesnumbered,ruled]{algorithm2e}
\usepackage{calc}
\usepackage{tikz}
\usepackage{booktabs}

\begin{document}

\maketitle

\input{text/00_Abstract}

\input{text/01_Introduction}
\input{text/02_RelatedWorks}
\input{text/03_Method}
\input{text/04_Experiments}
\input{text/05_Discussion}
\input{text/06_Conclusion}
\input{text/07_Acknowledgments}

\bibliography{references}
\end{document}


\maketitle



\section{Additional Details}
The pseudocode for our training algorithm is shown in \autoref{alg:training_algorithm}.
For our experiments, we use a neural network with $10$ layers and continuous levels of detail from $1.0$ up to $385.0$. The parameters for our network are laid out in \autoref{tab:model_parameters}.

\begin{algorithm}[!htbp]
 \KwData{Training images with poses}
 \KwResult{Trained LFN with continuous LODs}
 $lfn \leftarrow$ InitializeLFN()\\
 $optimizer\leftarrow$Adam($lfn$)\\
  \For{epoch $1$ \KwTo $num\_epochs$}{
    \For{$images \leftarrow$ GetImageBatch()}{
      $sat \leftarrow$ ComputeSAT($images$)\\
      $ray\_pdf \leftarrow$ ComputeRayPDF($images$)\\
      \For{$rays, colors \leftarrow$ SampleRays($ray\_pdf$)}{
        $low\_lod, low\_lod\_scale \leftarrow$ SampleLOD()\\
        $low\_lod\_colors \leftarrow$ SampleSAT($sat$, $rays$, $low\_lod\_scale$)\\
        $loss \leftarrow L2(lfn(rays, max\_lod), colors) + L2(lfn(rays, low\_lod), low\_lod\_colors)$\\
        $loss.backward()$\\
        $optimizer.step()$
      }
    }
 }
 \caption{Training Procedure Pseudocode for Continuous LOD LFNs}
 \label{alg:training_algorithm}
\end{algorithm}

\begin{algorithm}[!htbp]
 \KwData{Input feature $f$ from the variable-width linear layer and fractional LOD $\alpha$}
 \KwResult{Masked feature $f'$}
 $f' \leftarrow$ cat($f[:,:-1], \alpha * f[:,-1:]$, dim=-1)
 \caption{Neuron Masking Pseudocode for Continuous LOD LFNs}
 \label{alg:training_algorithm}
\end{algorithm}

\begin{table}[!htbp]
  \caption{Model Parameters for Each Level of Detail.}
  \label{tab:model_parameters}
  \centering
  \scalebox{0.95}{
  \begin{tabular}{lrrr}
    \toprule
    Level of Detail & 1.0 & $\ell$ & 385.0\\
    \midrule
    Model Layers & 10 & 10 & 10\\
    Layer Width  & 128 & $127 + \lceil \ell \rceil$ & 512\\
    Parameters   & 135,812 & $\approx 9 * (127 + \ell)^2$ & 2,116,100\\
    Model Size (MB)    & 0.518 & $\approx 36 * (127 + \ell)^2 / 2^{20}$ & 8.072\\
    Target Scale    & $1/8$ & $2\hat{\;}\left(4(\frac{127 + \ell}{512})-4\right)$ & $1$\\
    \bottomrule
  \end{tabular}
  }
\end{table}

\section{Additional Results}
We present some qualitative results in \autoref{fig:lods_results}. Additional qualitative results are available on our supplementary webpage.

\newcommand{\datasetlods}[2]{
    \begin{subfigure}[!htbp]{\linewidth}
      \centering
      \includegraphics[width=\linewidth]{figures/#2/lods.pdf}
      \caption{Dataset #1 LODs shown at various scales}
      \label{fig:qualitative_lods_vs_#2}
    \end{subfigure}
    \begin{subfigure}[!htbp]{\linewidth}
      \centering
      \includegraphics[width=0.88\linewidth]{figures/#2/lods_same_scale.pdf}
      \caption{Dataset #1 LODs shown at the same scale}
      \label{fig:qualitative_lods_ss_#2}
    \end{subfigure}
}
\begin{figure*}[!tbhp]
    \centering
    \datasetlods{A}{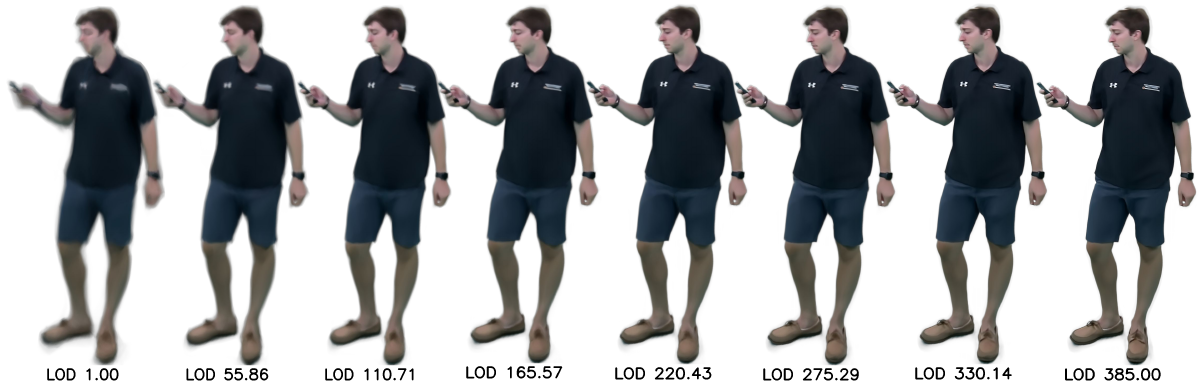}%
    \datasetlods{C}{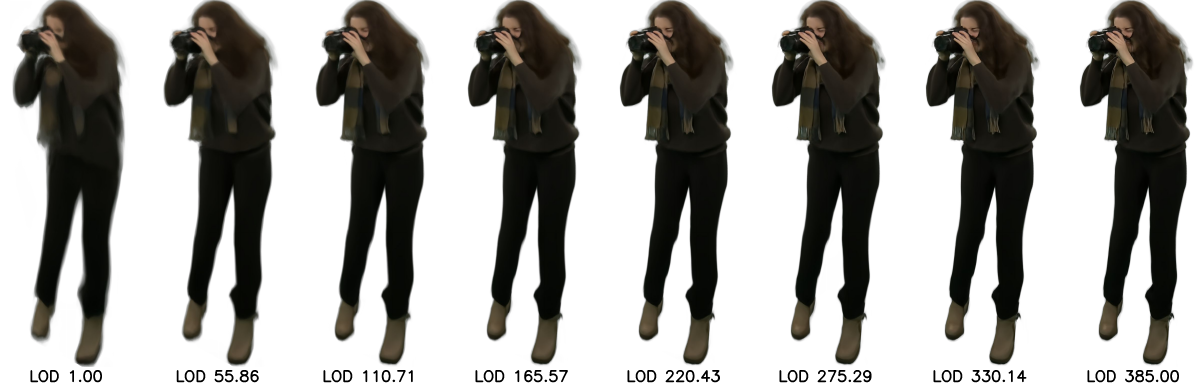}%
    \caption{Qualitative results rendering our continuous LFNs at multiple levels of detail for two datasets.}
    \label{fig:lods_results}
\end{figure*}

\subsection{Comparision to NeRF}
Neural radiance fields use volume rendering and 3D scene coordinates which provide 3D scene structure and multi-view consistency at the cost of requiring dozens to hundreds of evaluations per ray.
Two continuous LOD methods for NeRFs are Mip-NeRF~\cite{barron2021mipnerf} and Zip-NeRF~\cite{barron2023zipnerf}. 
Mip-NeRF uses integrated positional encoding to approximate a canonical frustum around a ray while Zip-NeRF uses multisampling of feature grid.
Both of these methods are targeted solely toward anti-aliasing and flicker reduction rather than towards resource adaptivity.
Hence, the entire model must be downloaded for rendering and the performance per pixel is the same at each scale.
Furthermore, neither method is directly applicable to light field networks which rely on the spectral bias of ReLU MLPs and thus are incompatible with positional encoding and feature grids.

For reference purposes, we present quantitative results using Mip-NeRF~\cite{barron2021mipnerf} to display our datasets in \autoref{tab:quantiative_quality_mipnerf}.
We train Mip-NeRF for $1$ million iterations with a batch size of $1024$ rays with the same $67\%$ foreground and $33\%$ background split in each batch.
We also use the same training and test split for each dataset as in our experiments.

\begin{table}[!htbp]
  \caption{Average Rendering Quality Comparison
  }
  \newcommand{\quantitativeQualityTabScale}{0.9}
  \label{tab:quantiative_quality_mipnerf}
  \centering
  \begin{subtable}[h]{\linewidth}
      \centering
      \scalebox{\quantitativeQualityTabScale}{
      \begin{tabular}{lrrrr}
          \hline
          Model & 1/8 & 1/4 & 1/2 & 1/1 \\ 
          \hline
Continuous LOD LFN & 28.06 & 29.79 & 28.44 & 27.40 \\
Mip-NeRF & 24.81 & 24.95 & 24.35 & 23.86 \\
          \hline
      \end{tabular}
      }
      \caption{PSNR (dB) at 1/8, 1/4, 1/2, and 1/1 scale.}
  \end{subtable}\\
  \vspace{2mm}
  \begin{subtable}[h]{\linewidth}
      \centering
      \scalebox{\quantitativeQualityTabScale}{
      \begin{tabular}{lrrrr}
          \hline
          Model & 1/8 & 1/4 & 1/2 & 1/1 \\ 
          \hline
Continuous LOD LFN & 0.8380 & 0.8751 & 0.8487 & 0.8455 \\
Mip-NeRF & 0.6819 & 0.6735 & 0.6451 & 0.6374 \\
          \hline
      \end{tabular}
      }
      \caption{SSIM at 1/8, 1/4, 1/2, and 1/1 scale.}
  \end{subtable}
\end{table}

\begin{figure}[!htbp]
    \centering
    \includegraphics[width=0.5\linewidth]{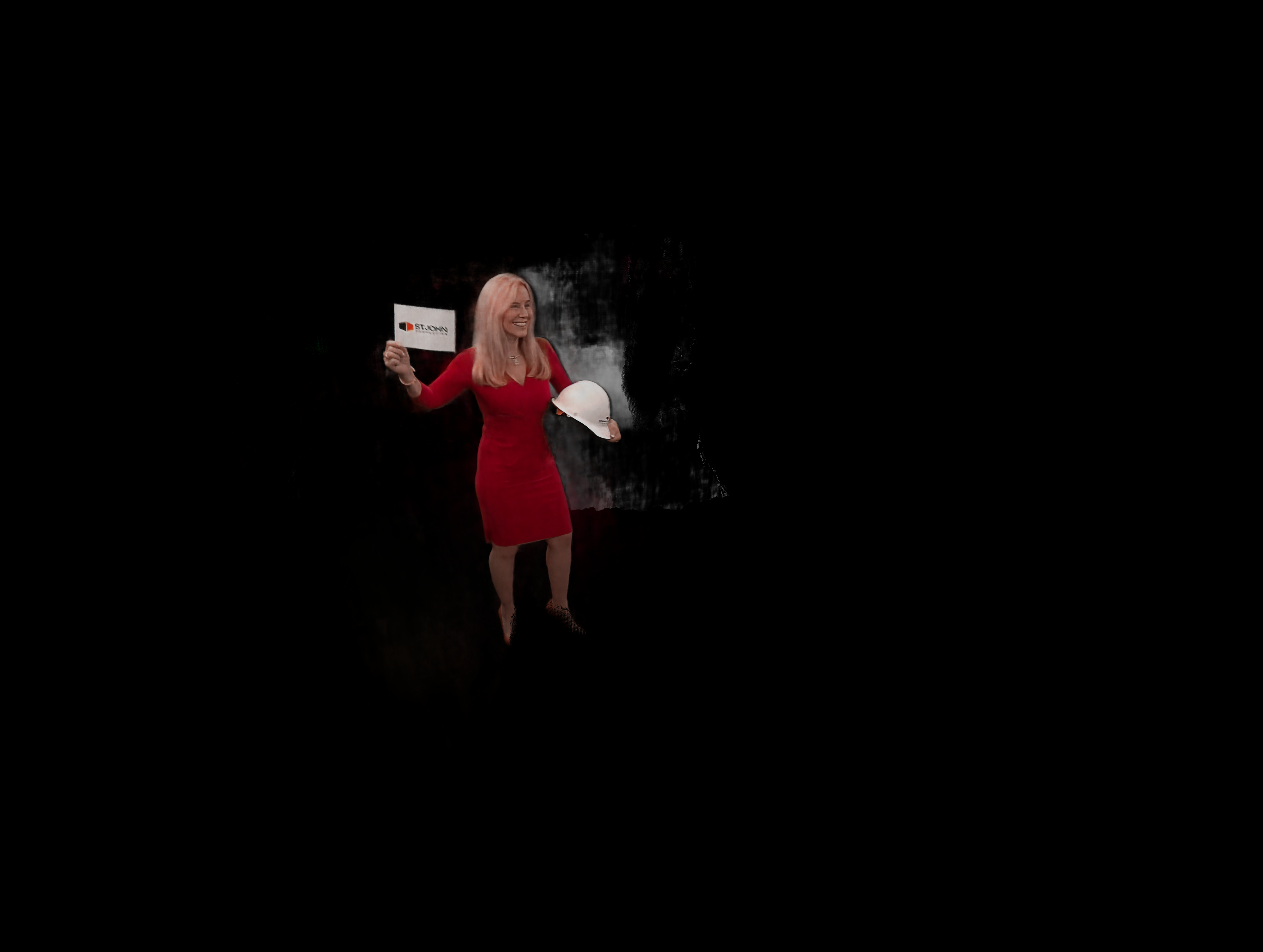}
    \vspace{1mm}
    \caption{Mip-NeRF rendering.}
    \label{fig:mipnerf_rendering}
\end{figure}

In our experiments, we observe that with our sampling scheme, Mip-NeRF is not able to separate the foreground and background cleanly as shown in \autoref{fig:mipnerf_rendering} which leads to worse PSNR and SSIM results.

In general, NeRF-based methods are better able to perform view-synthesis with high-frequency details due to their use of positional encoding and their 3D structure.
MLP-based methods such as Mip-NeRF typically have a compact size ($\leq 10$ MB) but suffer from slow rendering times on the order of tens of seconds per image.
Feature-grid NeRFs such as Instant-NGP~\cite{mueller2022instant}, Plenoxels~\cite{yu2021plenoxels}, and Zip-NeRF~\cite{barron2023zipnerf} can achieve real-time rendering but at the cost of larger model sizes ($\geq 30$ MB).
Factorized feature grids such as TensorRF~\cite{chen2022tensorf} promise both fast rendering and small model sizes.
Note that the goal of our paper is to enable more granularity with continuous levels of detail for rendering and streaming purposes rather than improving view-synthesis quality.

\bibliography{references}

%% file: text/00_Macros.tex
\usepackage{color}
\definecolor{ToRephaseColor}{rgb}{1.0, 0.4, 0.0}
\definecolor{TodoColor}{rgb}{1.0, 0.0, 1.0}
\definecolor{GreenColor}{rgb}{0.0, 0.0, 1.0}
\definecolor{PurpleColor}{rgb}{0.5, 0.0, 0.5}
\definecolor{BlueColor}{rgb}{0.0, 0.0, 1.0}
\definecolor{BabyBlueColor}{rgb}{0.5, 0.5, 1}
\definecolor{RevisionFixedColor}{rgb}{0.6, 0.0, 0.4}
\definecolor{FinalColor}{rgb}{0.0, 0.0, 0.0}

\def\Vhrulefill{\leavevmode\leaders\hrule height 0.7ex depth \dimexpr0.4pt-0.7ex\hfill\kern0pt}

%% file: text/00_Abstract.tex
\begin{abstract}
Recently, several approaches have emerged for generating neural representations with multiple levels of detail (LODs).
LODs can improve the rendering by using lower resolutions and smaller model sizes when appropriate.
However, existing methods generally focus on a few discrete LODs which suffer from aliasing and flicker artifacts as details are changed and limit their granularity for adapting to resource limitations.
In this paper, we propose a method to encode light field networks with continuous LODs, allowing for finely tuned adaptations to rendering conditions.
Our training procedure uses summed-area table filtering allowing efficient and continuous filtering at various LODs. Furthermore, we use saliency-based importance sampling which enables our light field networks to distribute their capacity, particularly limited at lower LODs, towards representing the details viewers are most likely to focus on.
Incorporating continuous LODs into neural representations enables progressive streaming of neural representations, decreasing the latency and resource utilization for rendering.
\end{abstract}

%% file: text/01_Introduction.tex
\section{Introduction}

\begin{figure}[!htbp]
  \includegraphics[width=\textwidth]{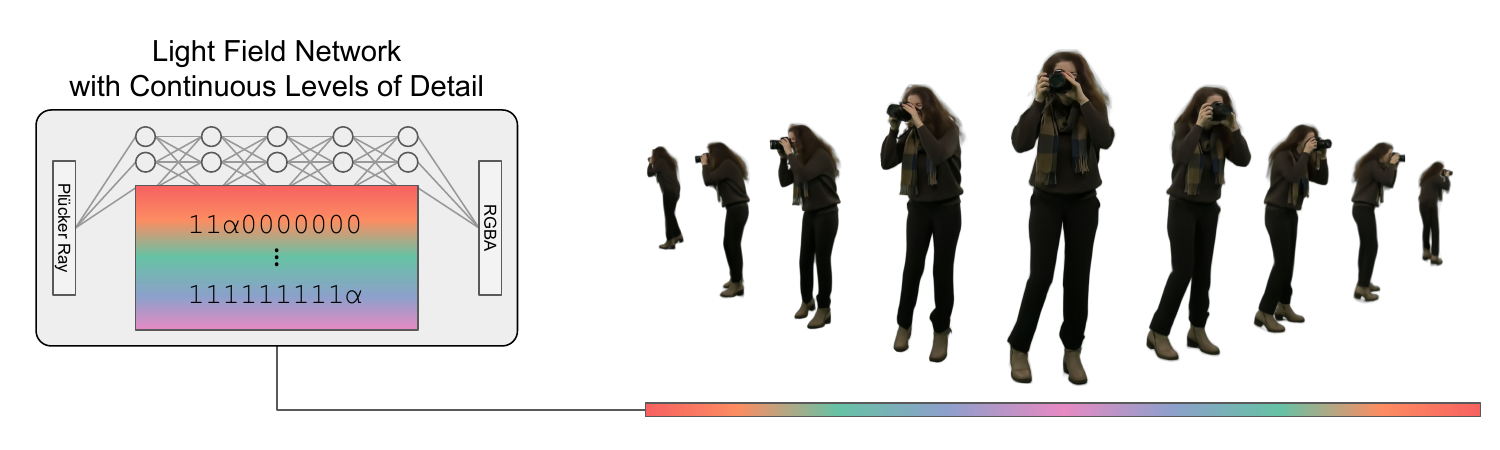}
  \caption{Our light field network features continuous levels of detail, enabled by training with summed-area table filtering and saliency-based importance sampling. Continuous levels of detail enable the interactive streaming of light fields with smooth transitions and finely tuned adaptivity.}
  \label{fig:teaser}
\end{figure}

In the past few years, implicit neural representations~\cite{mildenhall2020nerf,park2019deepsdf} have become a popular technique in computer graphics and vision for representing high-dimensional data such as 3D shapes with signed distance fields and 3D scenes captured from multi-view cameras. Light Field Networks (LFN)~\cite{sitzmann2021lfns} are able to represent 3D scenes with support for real-time rendering as each pixel of a rendered image only requires a single evaluation through the neural network.

In computer graphics, levels of detail (LODs) are commonly used to optimize the rendering process by reducing resource utilization for smaller distant objects in a scene. LODs prioritize resources to improve the overall rendering performance. In streaming scenarios, LODs can prioritize and reduce network bandwidth usage.
While LODs for implicit neural representations are beginning to be explored~\cite{cho2022streamable,chen2021multiresolution,Landgraf2022PINs,lindell2021bacon,li2022progressive}, most existing work focuses on offering a few discrete LODs which have three drawbacks for streaming scenarios. 
First, with only a few LODs, switching between them can result in flicker or popping effects as there are significant jumps from one LOD to the next. 
Second, discrete LODs require streaming in larger model deltas which take longer and create spikes in network activity. 
Third, transitioning across successive LODs can require rendering multiple levels and impact the rendering performance.
Continuous LODs resolve these challenges by allowing smoother transitions with finer quality adaptation and size differences across LODs.
Additionally, they allow rendering engines to dynamically adjust the rendering quality based on real-time bandwidth and compute resource availability without needing the viewer to select the LOD.

In this paper, we develop a method to achieve continuous levels of detail for neural representations, focusing on light field networks. 
In summary, our light field networks with continuous LODs have the following benefits:
\begin{itemize}
    \item Continuous LODs allow us to smoothly transition across LODs without popping artifacts,
    \item In streaming scenarios, the LOD can be increased or decreased by downloading only one additional row and column of weights per layer, allowing lower latency transitions with smoother network patterns, and
    \item Importance sampling allows LFNs to focus their capacity on the most salient regions of the light field, allowing details to resolve at lower LODs.
\end{itemize}

%% file: text/02_RelatedWorks.tex
\section{Background and Related Works}
In this section, we provide some background on general neural fields and more specifically light field networks. We then overview recent methods for achieving multiple levels of detail with neural fields.
For a general overview of neural rendering, we refer readers to Tewari~\etal~\cite{tewari2021advances}.

\subsection{Implicit Neural Representations}
Coordinate-based neural networks have been used to encode various signals such as images~\cite{sitzmann2019siren}, signed distance functions~\cite{park2019deepsdf}, radiance fields~\cite{mildenhall2020nerf,barron2021mipnerf,barron2023zipnerf}, and light fields~\cite{feng2021signet,sitzmann2021lfns}.
These neural network-based models are often referred to as implicit neural representations or neural fields.
Among these representations, neural radiance fields (NeRFs) and light field networks (LFNs) are both able to represent colored 3D scenes with view-dependant appearance effects.

Neural radiance fields (NeRFs)~\cite{mildenhall2020nerf} employ differentiable volume rendering to encode a 3D scene into a multi-layer perceptron (MLP) neural network. 
By learning the density and color of the scene and using a positional encoding, NeRF can perform high-quality view synthesis, rendering the scene from arbitrary camera positions, while maintaining a very compact representation.
However, the original NeRF implementation has many drawbacks, such as slow rendering times, which has limited its practicality. 
With an incredible amount of interest in neural rendering, many follow-up works have been proposed to improve NeRFs with better rendering performance~\cite{yu2021plenoxels,reiser2021kilonerf,deng2022fovnerf}, better quality~\cite{barron2021mipnerf}, generalizability~\cite{gu2021stylenerf}, and deformations~\cite{park2021hypernerf,pumarola2020d,park2021nerfies}.
Additionally, feature grid methods~\cite{mueller2022instant, yu2021plenoxels} enable learning scenes in seconds and rendering in real-time.
Importance sampling~\cite{zhang2022fast} can achieve faster learning with fewer training rays.

\paragraph{Light Field Networks}

\begin{figure}[!htbp]
    \centering
    \includegraphics[width=0.9\linewidth]{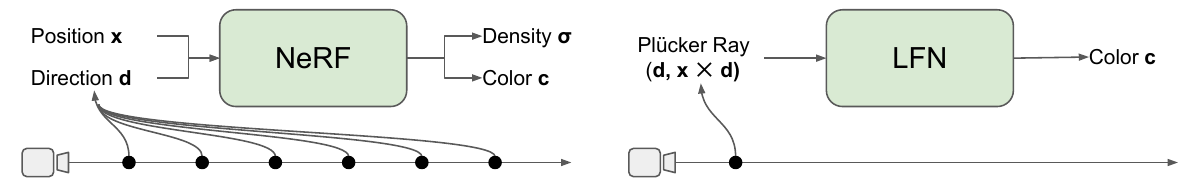}
    \caption{LFNs directly predict the RGB color for each ray in a single inference using Pl\"{u}cker coordinates, avoiding the dozens to hundreds of inferences required by NeRFs.}
    \label{fig:illustration_nerf_vs_lfn}
\end{figure}

Light Field Networks (LFNs)~\cite{sitzmann2021lfns,feng2021signet,li2022neulf,chandramouli2021light,cao2022realtimeneulf} encode light fields~\cite{levoy1996lightfieldrendering,gortler1996lumigraph} by directly learning the 4D variant of the plenoptic function for a scene.
Specifically, LFNs directly predict the emitted color for a ray which eliminates the need for volume rendering, making light fields much faster to render compared to other neural fields.
Earlier work in light field networks focus on forward-facing scenes using the common two-plane parameterization for light fields.
SIGNET~\cite{feng2021signet,feng2022subspaces} uses Gegenbauer polynomials to encode light field images and videos.
NeuLF~\cite{li2022neulf} proposes adding a depth branch to encode light fields from a sparser set of images.
Pl\"{u}cker coordinates have been used~\cite{sitzmann2021lfns, feng2022prif}  to represent 360-degree light fields.

\subsection{Levels of Detail}
Several methods have been proposed for neural representations with multiple levels of detail.
NGLOD~\cite{takikawa2021neural} encode signed distance functions into a multi-resolution octree of feature vectors. 
VQAD~\cite{takikawa2022variable} adds vector quantization with a feature codebook and presents results on NeRFs.
BACON~\cite{lindell2021bacon} encodes LODs with different Fourier spectrums for images and radiance fields.
PINs~\cite{Landgraf2022PINs} develop a progressive Fourier feature encoding to improve reconstruction and provide progressive LODs.
MINER~\cite{Vishwanath2022miner} trains neural networks to learn regions within each scale of a Laplacian pyramid representation.
Streamable Neural Fields~\cite{cho2022streamable} propose growing neural networks to represent increasing spectral, spatial, or temporal sizes.
Progressive Multi-Scale Light Field Networks~\cite{li2022progressive} train a light field network to encode light fields at multiple resolutions.

To generate arbitrary intermediate LODs, existing methods blend outputs across discrete LODs. With only a few LODs, the performance does not scale smoothly since the next discrete LOD must be computed entirely. Our method offers continuous LODs with hundreds of performance levels allowing for finer adaptation to resource limitations.

%% file: text/03_Method.tex
\section{Method}
Our method primarily builds upon \textit{Light Field Networks} (LFNs)~\cite{sitzmann2021lfns}.
Specifically, we represent rays $\mathbf{r}$ in Pl\"{u}cker coordinates $(\mathbf{r}_d, \mathbf{r}_o \times \mathbf{r}_{d})$ which are input to a multi-layer perceptron (MLP) neural network without any positional encoding.
The MLP directly predicts RGBA color values without any volume rendering or other accumulation.
Each light field network is trained to overfit a single static scene.

\subsection{Arbitrary-scale Arbitrary-position Sampling with Summed Area Tables}

\begin{figure}[!htbp]
    \centering
    \hfill
     \begin{subfigure}[b]{0.45\linewidth}
         \centering
         \includegraphics[width=0.8\linewidth]{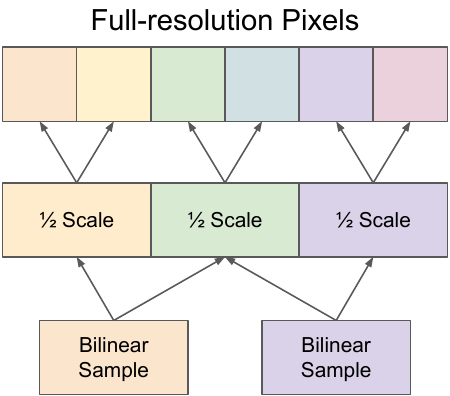}
         \caption{Discrete Scale Sampling}
     \end{subfigure}\hfill
     \begin{subfigure}[b]{0.45\linewidth}
         \centering
         \includegraphics[width=0.8\linewidth]{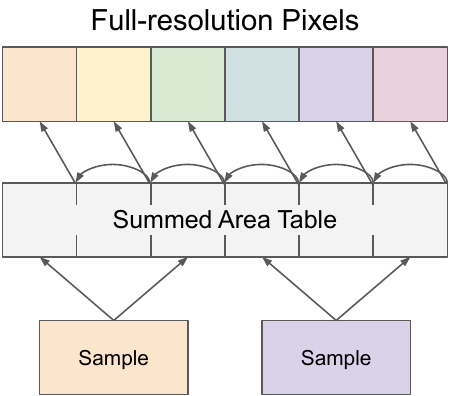}
         \caption{Summed Area Table Sampling}
     \end{subfigure}
     \hfill{}
    \caption{An illustration of discrete and summed-area table sampling. (a) Sampling from a discrete resolution requires linear interpolation from a downsampled image to the target scale and position. (b) Summed area tables allow us to sample at both arbitrary scales and positions without significant additional memory or compute.}
    \label{fig:mipmap_vs_sat_sampling_illustration}
\end{figure}

In order to reduce aliasing and flickering artifacts when rendering at smaller resolutions, e.g. when an object is far away from the user, lower levels of details need to be processed with filtering to the appropriate resolution.
In prior work, multi-scale LFNs~\cite{li2022progressive} are trained on images resized to $1/2$, $1/4$, and $1/8$ scale using area downsampling. During training, rays are sampled from the full-resolution image while colors are sampled from lower-resolution images using bilinear sampling. While training on lower-resolution light fields yields multi-scale light field networks, the bilinear subsampling of the light field may not provide accurate filtered colors for intermediate positions. 
As shown in \autoref{fig:mipmap_vs_sat_sampling_illustration}, colors for higher-resolution rays get averaged over a larger area when performing bilinear subsampling in between low-resolution pixels.

Another method for generating multi-scale light fields is to apply to filter at full resolution to get a spatially accurate anti-aliased sample for each pixel location. Naively precomputing and caching full-resolution copies of each light field image at each scale would significantly increase memory usage. Computing the average pixel color for each sampled ray at training time would require additional computation.
Summed area tables~\cite{crow1984sat, li2021logrectilinear} can be used to efficiently sample pixels at arbitrary scales and positions, allowing us to sample from filtered versions of the training image without caching multiple copies. Sampling from a summed area table is a constant time operation, giving us an average over any axis-aligned rectangular region with only four samples.
With additional samples, summed-area tables can also be used to apply higher-order polynomial (e.g. cubic) filters~\cite{heckbert1986filtering,hensley2005fast} or Gaussian filters~\cite{kovesi2010gaussian} for even better anti-aliasing, though we only use box filtering in our implementation.

\subsection{Continuous Levels of Detail}

\begin{figure}[!htbp]
    \centering
    \hfill
    \begin{subfigure}[!t]{0.45\linewidth}
        \centering
        \includegraphics[width=\linewidth]{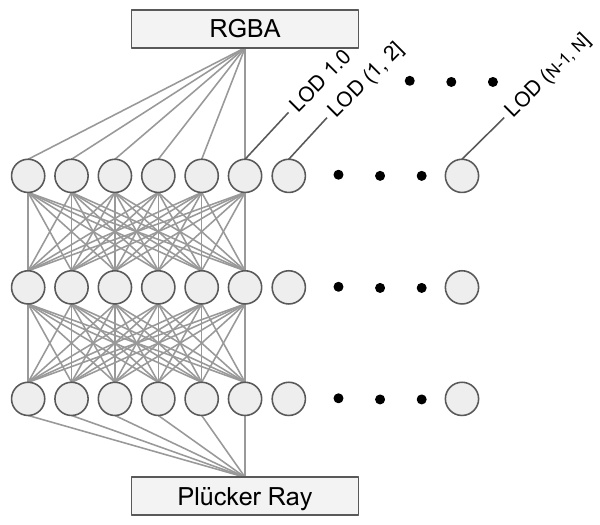}
        \caption{Variable-size layers. By assigning a level of detail to every network width, we can achieve hundreds of performance levels.}
        \label{fig:continuous_lod_illustration}
    \end{subfigure}\hfill%
    \begin{subfigure}[!t]{0.45\textwidth}
        \centering
        \includegraphics[width=\linewidth]{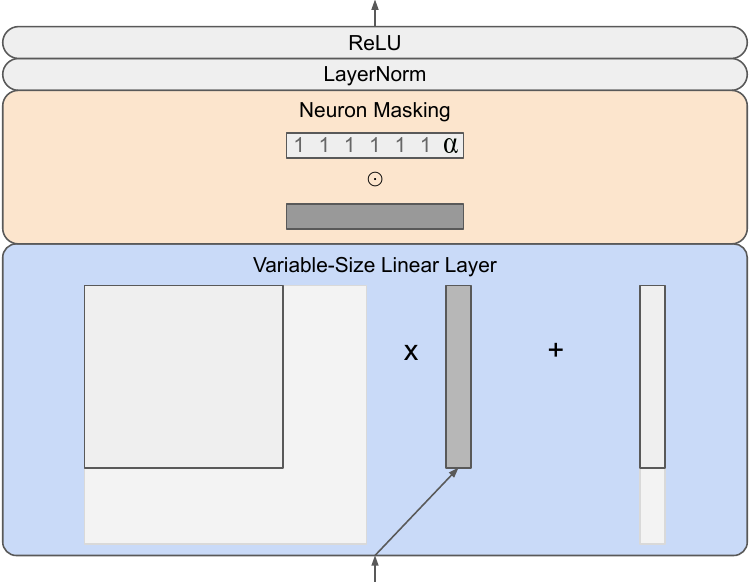}
        \caption{Neuron masking. To render at continuous LOD $\ell$, we use weights corresponding to LOD $\lceil \ell \rceil$ and apply alpha-blending to continuously fade-in features produced by the new neuron with $\alpha = \ell - \lfloor \ell \rfloor$ at each layer.}
        \label{fig:neuron_masking_illustration}
    \end{subfigure}
    \hfill{}
    \caption{Illustrations of our method to achieve continuous levels of detail.}
    \label{fig:continuous_lod_methods}
\end{figure}

While previous neural field methods offer static levels of detail corresponding to fixed scales~\cite{li2022progressive,takikawa2022variable,chen2021multiresolution} or fixed spectral frequency bands~\cite{lindell2021bacon,cho2022streamable}, our goal is to generate a finer progression with continuous levels of detail. Continuous levels of detail enable smoother transitions and more precise adaptation to resolution and resource requirements.

Following existing work~\cite{cho2022streamable, li2022progressive, yu2019universally}, we encode levels of detail using different widths of a single multi-layer perception neural network.
Unlike Mip-NeRF~\cite{barron2021mipnerf,barron2022mipnerf360}, this enables optimized performance with smaller neural networks at lower levels of detail.
However, for continuous levels of detail, we propose two changes.
First, we map the desired level of detail to every available width to extend a few levels of detail to hundreds of levels of detail as shown in \autoref{fig:continuous_lod_illustration}.
Second, we propose neuron masking which fades in new neurons to enable true continuous quality adjustments.

\paragraph{LOD to Scale Mapping}
Li~\etal~\cite{li2022progressive} train multi-scale LFNs which use width factors $1/4$, $2/4$, $3/4$, and $4/4$ (128, 256, 384, 512 widths) to encode $1/8$, $1/4$, $1/2$, and $1/1$ scale light fields respectively. To extend this to arbitrary widths, we formulate the following equations which describe the correspondence between network width $w$ and light field scale $s$:
\begin{align}
    s &= 2\hat{\;\;}(4w-4) \\
    w &= (1/4)*(\log_{2}(s) + 4)
\end{align}
By using the above equations, we can assign a unique scale to each width sub-network in our multi-scale light field network. Since this is a one-to-one invertible mapping, we can also compute the ideal level of detail to use for rendering at any arbitrary resolution. In our experiments, we use a minimum width of $25\%$ of nodes corresponding to a scale of $1/8$ to ensure a reasonable minimum quality and training image size. As an example, for a network with $512$-width hidden layers, the lowest level of detail uses only $128$ neurons of each hidden layer while the highest uses $512$.

\paragraph{Neuron Masking}
Since neural networks have discrete widths, it is necessary to map continuous levels of detail to discrete widths.
Hence, we propose to use neuron masking to provide true continuous levels of detail with discrete-sized neural networks.
As weights corresponding to each new width become available, we propose to apply alpha-blending on neurons corresponding to the width.
This alpha-blending enables features from existing neurons to continuously transition, representing any intermediate level of detail between the discrete widths.
Given feature $\mathbf{f}$ and fractional LOD $\alpha = l - \lfloor l \rfloor$, the new feature $\mathbf{f}'$ with neuron masking is the element-wise product:
\begin{equation}
\mathbf{f}' = (1, ..., 1, \alpha)^\top \odot \mathbf{f}
\end{equation}

\subsection{Saliency-based Importance Sampling}

With continuous LODs representing light fields at various scales, the capacity of the LFN is constrained at lower LODs. Hence, details such as facial features may only resolve at higher levels of detail.
To maximize the apparent fidelity, the capacity of the network should be distributed towards the most salient regions, {\em i.e.} the areas where viewers are most likely to focus.
We propose to use saliency-based importance sampling which focuses training on salient regions of the light field.
For all foreground pixels, we assign a base sampling weight $\lambda_{f}$ and add a weight of $\lambda_{s}*s$ based on the pixel saliency $s$.
Specifically, for a given foreground pixel $x$ in a training image with saliency $s$, we sample from the probability density:
\begin{equation}
    p(x) = \lambda_{f} + \lambda_{s}*s
\end{equation}

In our experiments, we use $(\lambda_{f}, \lambda_s) = (0.4, 0.6)$ which yields reasonable results. 
At each iteration, we sample $67\%$ of rays in each batch from foreground pixels using the above density. The remaining $33\%$ of rays are uniformly sampled from background pixels.

\newcommand{\datasetlods}[2]{
    \begin{subfigure}[!htbp]{\linewidth}
      \centering
      \includegraphics[width=\linewidth]{figures/#2/lods.pdf}
      \caption{Dataset #1 LODs shown at various scales}
      \label{fig:qualitative_lods_vs_#2}
    \end{subfigure}
    \begin{subfigure}[!htbp]{\linewidth}
      \centering
      \includegraphics[width=0.88\linewidth]{figures/#2/lods_same_scale.pdf}
      \caption{Dataset #1 LODs shown at the same scale}
      \label{fig:qualitative_lods_ss_#2}
    \end{subfigure}
}

%% file: text/04_Experiments.tex
\section{Experiments}
We conduct several experiments to evaluate whether our light field networks with continuous LODs overcome the problems with discrete LODs. We also conduct quality and performance evaluations to determine the compute and bandwidth overhead associated with continuous LODs.

\subsection{Experimental Setup}
We conduct our experiments using five light field datasets. 
Scenes are captured using $240$ cameras with $40\times6$ layout around the scene and a $4032\times3040$ resolution per camera.
Each dataset includes camera parameters extracted using COLMAP~\cite{schoenberger2016sfm,schoenberger2016mvs} and is processed with background matting. Of the 240 images, we use 216 for training, 12 for validation, and 12 for testing.
We generate saliency maps using the mit1003 pretrained network\footnote{From \url{https://github.com/alexanderkroner/saliency}} of Kroner~\etal~\cite{kroner2020contextual}.

For our model, we use an MLP with nine hidden layers and one output layer. Each hidden layer uses LayerNorm and ReLU. We use a minimum width of $128$ and a maximum width of $512$ for variable-size layers. 
Our models are trained using a squared L2 loss for the RGBA color with $8192$ rays per batch. In all of our experiments, we train using the Adam optimizer with the learning rate set to $0.001$ and exponentially decayed by $\gamma=0.98$ after each epoch. We train for 100 epochs. 
Each of our models is trained using a single NVIDIA RTX 2080 Ti GPU. 
Our PyTorch implementation and processed datasets are available at \url{https://augmentariumlab.github.io/continuous-lfn/}.

\subsection{Ablation Experiments}

\newcommand{\datasetSaliencyAblation}[2]{
    \begin{subfigure}[!htbp]{0.48\linewidth}
      \centering%
      \includegraphics[width=\linewidth]{figures/#2/saliency_ablation_a0.pdf}%
      \label{fig:qualitative_sal_ablation_#2}%
    \end{subfigure}%
}
\begin{figure}[!tbhp]
\begin{minipage}{0.49\linewidth}
    \centering\hfill%
    \datasetSaliencyAblation{A}{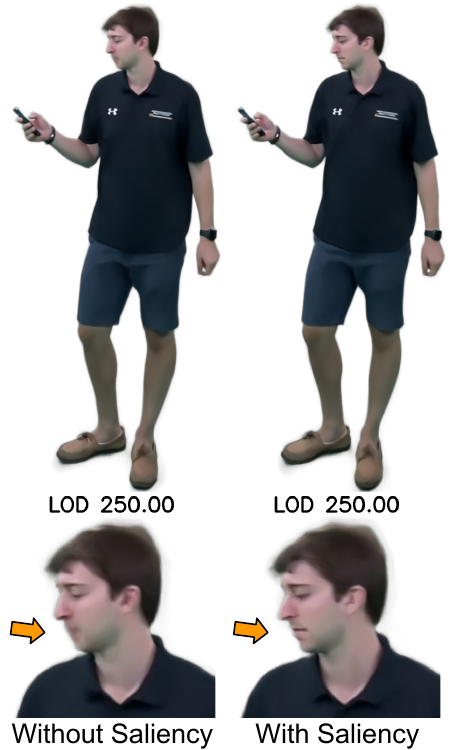}\hfill%
    \datasetSaliencyAblation{D}{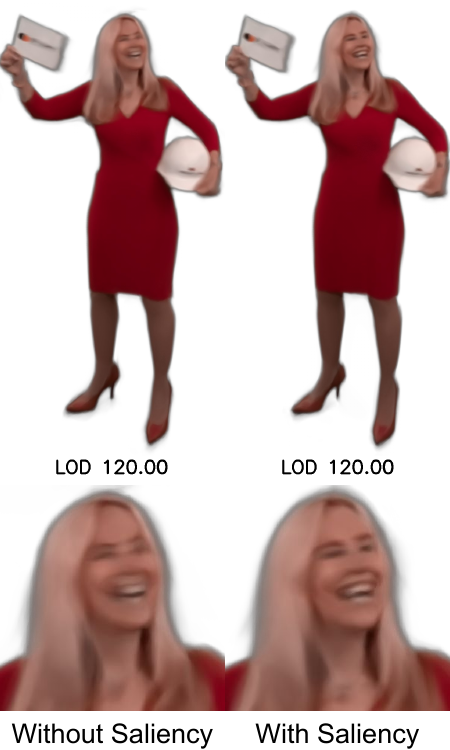}\hfill%
    \caption{Lower LOD ablation results show the effects of our saliency-based importance sampling. With importance sampling, features such as eyes and mouths in the salient regions resolve at earlier, lower LODs.}
    \label{fig:saliency_ablation_qualitative_results}
\end{minipage}\hfill%
\begin{minipage}{0.49\linewidth}
    \centering
    \includegraphics[width=\linewidth]{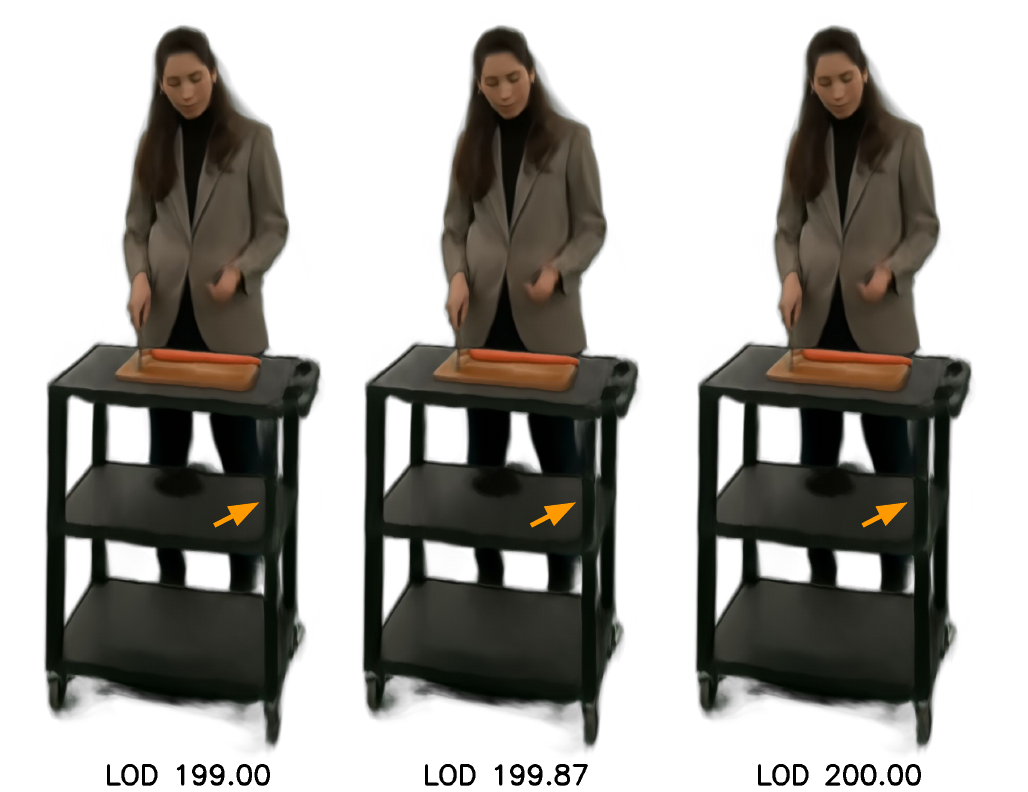}
    \caption{Neuron masking enables generating continuous levels of detail from discrete model widths. In the figure above, we see that one leg of the cart expands as the fractional level of detail $\alpha$ fades in new neurons. This effect is better seen in the accompanying supplementary video.}
\label{fig:neuron_masking_example}
\end{minipage}
\end{figure}

Our ablation experiments evaluate how each aspect of our method affects the final rendered quality. 
First, we replace the discrete resolution sampling in discrete-scale light field networks~\cite{li2022progressive} with our summed area table sampling.
Next, we add continuous LODs training which is enabled by arbitrary-scale filtering with summed-area tables.
Finally, we compare the prior two setups with our full method which also includes saliency-based importance sampling.

\subsection{Transitions across LODs}

\begin{figure}[!htbp]
    \centering
    \hfill
    \begin{subfigure}[!htbp]{0.5\linewidth}
        \centering
        \includegraphics[width=\linewidth]{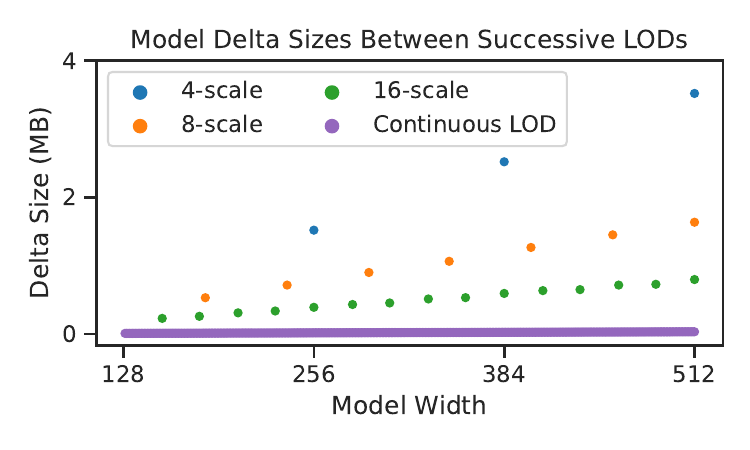}
        \caption{Model Delta Sizes}
        \label{fig:model_delta_sizes}
    \end{subfigure}\hfill%
    \begin{subfigure}[!htbp]{0.5\linewidth}
        \centering
        \includegraphics[width=\linewidth]{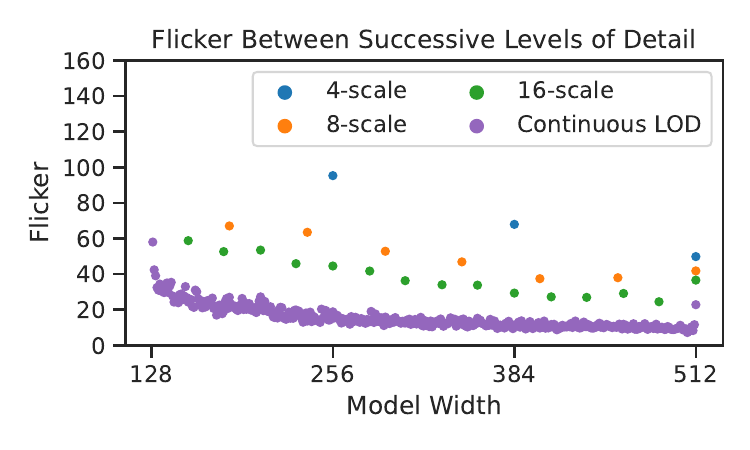}
        \caption{Average Flicker}
        \label{fig:flicker_plot}
    \end{subfigure}\hfill{}
    \caption{Plots showing the effects of transitioning across LODs. Transitioning with discrete LODs leads to larger network traffic spikes and more flickering.}
    \label{fig:quantitative_transition_plots}
    \vspace{-4mm}
\end{figure}

With continuous LODs, our method allows smooth transitions across LODs as additional bytes are streamed over the network or as the viewer approaches the subject. To quantitatively evaluate the smoothness of the transitions, we use the reference-based temporal flicker metric of Winkler~\etal~\cite{winkler2004perceptual}. This flicker metric first computes the difference $d$ between the processed images and reference images for two consecutive frames. Next, a difference image $c=d_{n}-d_{n-1}$ is computed across consecutive frames. The 2D discrete Fourier transform of the image $c$ is computed and values are summed based on the radial frequency spectrum into low and high-frequency sums: $s_L$ and $s_H$. Finally, the flicker metric is computed by adding these together: $\text{Flicker}=s_L + s_H$.

We compare against three discrete-scale baselines with 4, 8, and 16 levels of detail, with 8 and 16 LODs trained using summed-area table sampling. In our continuous LOD case, we render views at the highest LOD corresponding to each discrete width (i.e. LOD $1.0$, $2.0$, ..., $385.0$), using the static ground truth view as the reference frames. Flicker values are computed for each LOD using the transition from the next lower LOD and then averaged across all test views.
Our flicker results are shown in \autoref{fig:flicker_plot}.
With only four LODs, the discrete-scale LFN method has three transitions, each with large model deltas (up to $3.5$ MB) and high flicker values.
Additional levels of detail reduce the model delta sizes and the flicker values with our continuous LOD method minimizing the model delta sizes and the flicker values.
With our method, the LOD can be transitioned in small ($\leq32$ KB) gradual steps.

\subsection{Rendering Quality}

\begin{table}[!htbp]
    \centering
     \begin{subtable}[b]{0.5\linewidth}
         \centering
         \scalebox{0.73}{
        \begin{tabular}{lcccc}
             \toprule
             Model & 1/8 & 1/4 & 1/2 & 1/1 \\
             \midrule
Discrete-scale LFN & 29.34 & 29.68 & 28.39 & 27.41 \\
+ SAT filtering & 29.77 & 29.99 & 28.54 & 27.46 \\
+ Continuous LOD & 27.87 & 29.77 & 28.38 & 27.38 \\
+ Importance Sampling & 28.06 & 29.79 & 28.44 & 27.40 \\
             \bottomrule
        \end{tabular}
         }
         \caption{PSNR (dB)}
     \end{subtable}%
     \begin{subtable}[b]{0.5\linewidth}
         \centering
         \scalebox{0.73}{
        \begin{tabular}{lcccc}
             \toprule
             Model & 1/8 & 1/4 & 1/2 & 1/1 \\
             \midrule
Discrete-scale LFN & 0.8809 & 0.8763 & 0.8503 & 0.8465 \\
+ SAT filtering & 0.8898 & 0.8806 & 0.8513 & 0.8466 \\
+ Continuous LOD & 0.8370 & 0.8743 & 0.8484 & 0.8460 \\
+ Importance Sampling & 0.8380 & 0.8751 & 0.8487 & 0.8455 \\
             \bottomrule
        \end{tabular}
        }
         \caption{SSIM}
    \end{subtable}
    \caption{Quantitative Training Ablation Results at 1/8, 1/4, 1/2, and 1/1 scales. Each scale is evaluated at its corresponding LOD.}
    \label{tab:ablation_quality}
\end{table}

\begin{figure}[!htbp]
    \centering
    \hfill
    \begin{subfigure}[!htbp]{0.5\linewidth}
        \centering
        \includegraphics[width=\linewidth]{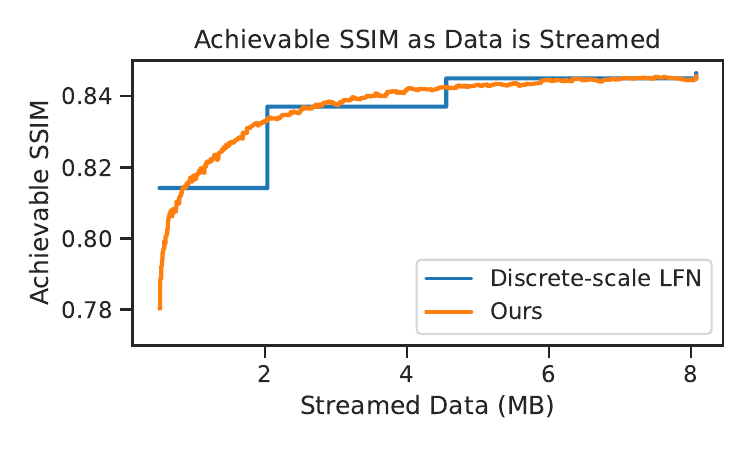}
        \caption{Average SSIM quality at the full resolution.}
        \label{fig:psnr_ssim_plot}
    \end{subfigure}\hfill%
    \begin{subfigure}[!htbp]{0.5\linewidth}
        \centering
        \includegraphics[width=\linewidth]{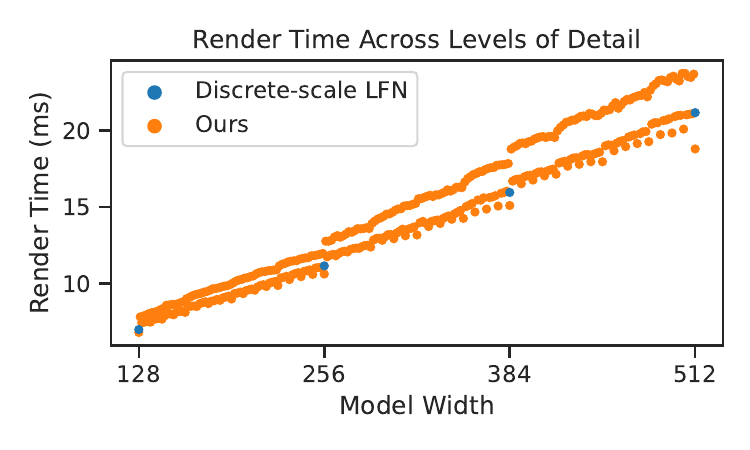}
        \caption{Average render times at $1/4$ scale.}
        \label{fig:render_time_plot}
    \end{subfigure}\hfill{}
    \caption{Plots showing our quantitative evaluation results. With continuous LODs, the LOD can be dynamically adjusted to maximize the quality based on available resources.}
    \label{fig:quantitative_plots}
\end{figure}

Quantitative PSNR and SSIM results are shown in \autoref{tab:ablation_quality}. 
First, we see that adding summed-area table filtering to discrete-scale light field networks with four scales results in slightly improved PSNR and SSIM results while enabling arbitrary-scale sampling. Training a continuous LOD network impacts the performance at the original four LODs but allows us to have continuous LODs. Adding importance sampling allows us to focus on salient regions without significantly impacting the quantitative results.

Qualitative results of our saliency-based importance sampling ablation are shown in \autoref{fig:saliency_ablation_qualitative_results}. We see that details along faces appear at earlier LODs when using saliency for importance sampling. All of these details resolve at the highest LODs with and without using importance sampling.

\subsection{Rendering Performance}

We evaluate the rendering performance by rendering training views across each LOD. For our rendering benchmarks, we use half-precision inference and skip empty rays with the auxiliary network which evaluates ray occupancy.
Rendering performance results across the LODs are shown in \autoref{fig:render_time_plot}. We observe that as the LOD increases according to the width of the neural network, rendering times increase as well.
When rendering from a discrete-scale light field network with only four LODs, the user or application would need to select either the next higher or lower LOD, compromising on either the performance or the quality.
With continuous LODs, software incorporating our light field networks would be able to gradually increase or decrease the LOD to maintain a better balance between performance and quality.
In cases where the ideal model size is not known, continuous LODs allow dynamic adjusting of the LOD to satisfy a target frame rate.
In our PyTorch implementation, we observe that LODs with odd model widths have a slower render time than LODs with even model widths. LODs with model widths that are a multiple of eight perform slightly faster than other even model widths.

%% file: text/05_Discussion.tex
\section{Discussion}

By requiring light field networks to output reasonable results at each possible hidden layer width and incorporating neuron masking, we can achieve continuous of LODs. However, this applies additional constraints on the network as it needs to produce additional outputs. In our experiments, we observe slightly worse PSNR and SSIM results at the specific LODs corresponding to the $1/8$ and $1/4$ scales compared to the discrete-scale LFN which is trained with only four LODs. This is expected due to the additional constraints and less supervision at those specific LODs. The goal of our importance sampling procedure is to improve the quality of the salient regions of the light field rather than to maximize quantitative results.

Light field networks require additional cameras compared to neural radiance fields due to the lack of multi-view consistency prior provided by volume rendering. Hence, training light field networks requires additional cameras or  regularization~\cite{Feng2022Viinter} compared to NeRF methods. 
Furthermore, light field networks do not use positional encoding~\cite{tancik2020fourier} and represent high-frequency details as faithfully as NeRF methods.
As the primary goal of our work is to enable highly granular rendering trade-offs with more levels of detail, we leave these limitations to future work.

%% file: text/06_Conclusion.tex
\section{Conclusion}
In this paper, we introduce continuous levels of details for light field networks using three techniques. First, we introduce summed area table sampling to sample colors from arbitrary scales of an image without generating multiple versions of each training image in a light field. Second, we achieve continuous LODs by combining arbitrary-width networks with neuron masking. Third, we train using saliency-based importance sampling to help details in the salient regions of the light field resolve at earlier LODs. With our method for continuous LODs, we hope to make light field networks more practical for 6DoF desktop and virtual reality applications~\cite{du2019geollery,du2019projectgeollery,li2020meteovis}. 

%% file: text/07_Acknowledgments.tex
\section*{Acknowledgments}
We would like to thank Jon Heagerty, Sida Li, and Barbara Brawn for developing our light field datasets as well as the anonymous reviewers for the valuable comments on the manuscript.
This work has been supported in part by the NSF Grants 18-23321, 21-37229, and 22-35050 and the State of Maryland’s MPower initiative. 
Any opinions, findings, conclusions, or recommendations expressed in this article are those of the authors and do not necessarily reflect the views of the research sponsors.